\documentclass[letterpaper,10pt,conference]{ieeeconf}

\IEEEoverridecommandlockouts
\usepackage{amsmath}
\usepackage{amssymb}
\usepackage{booktabs}
\usepackage[table]{xcolor}
\usepackage{graphicx}
\usepackage{multirow}
\usepackage{cite}
\usepackage{placeins}
\usepackage{float}
\usepackage{afterpage}
\usepackage[hidelinks]{hyperref}

\title{\LARGE \bf
ProLoop: Progress-Aware Closed-Loop Framework\\
via Structured State Relations for Long-Horizon Robotic Task Planning
}

\author{Runxin Zou$^{1\dag}$, QingMiao Liang$^{1\dag}$, Jiatao Zhang$^{1}$, Xinyan Lv$^{1}$,
Liangyu Qie$^{1}$, Meiyi Bu$^{1}$, Wei Song$^{1\star}$, Shiqiang Zhu$^{1\star}$
\thanks{$^{1}$Zhejiang University, Hangzhou, China}%
\thanks{$^{\star}$Corresponding authors: Wei Song and Shiqiang Zhu.}%
\thanks{$^{\dag}$ Contribute equally to this work.}
}

\begin{document}
\bstctlcite{ProLoopBSTcontrol}

\maketitle
% arXiv preprint: enable page numbers on every page, including the title page
% (\maketitle otherwise applies an empty-footer title page style).
\thispagestyle{plain}
\pagestyle{plain}

\begin{abstract}
Long-horizon robotic tasks are commonly decomposed into semantic subgoals
to simplify planning. However, subgoal progress in LLM-based planning is
typically inferred implicitly from interaction histories. Inaccurate estimates
can misguide execution and allow local errors to propagate.
To address this, we propose \textbf{ProLoop}, a Progress-Aware Closed-Loop
framework that derives explicit subgoal progress from structured relations between
desired and execution states. A learned State-Relation Monitor assesses these
relations, which are then aggregated into subgoal progress as
ongoing, deviated, or completed. The resulting progress guides execution
adaptation, enabling reliable progress-aware control throughout execution.
Experiments on AgentBoard ALFWorld and a real-world robotic platform show that
ProLoop improves success rate by 12.94\% overall and by
15.46\% on Hard tasks, while reducing environment interactions
by 32.3\%.
\end{abstract}

%%%%%%%%%%%%%%%%%%%%%%%%%%%%%%%%%%%%%%%%%%%%%%%%%%%%%%%%%%%%%%%%%%%%%%%%%%%%%%%%
\section{Introduction}

Recent advances in robot task planning increasingly leverage large language models (LLMs) to
generate coherent multi-step plans
\cite{singh2023progprompt,zhang2024fltrnn}.
Despite strong performance on complex tasks, LLM-based planning remains susceptible to hallucinations
and logical inconsistencies
\cite{huang2024planning,chakraborty2025heal}.
To manage long-horizon planning complexity, hierarchical methods
decompose tasks into semantic subgoals, allowing each stage to focus on a simpler intermediate objective
\cite{ajay2023hip,liu2025delta,hu2025hiagent}.
However, during subgoal execution, subgoal progress is still commonly inferred implicitly by the LLM
from accumulated interaction trajectories and multi-source information
\cite{kim-hwang-2025-coex}.
Inaccurate progress estimates can therefore misguide execution, allowing local errors to accumulate
and propagate across subsequent subgoals.

\begin{figure}[t!]
    \centering
    \includegraphics[width=\columnwidth]{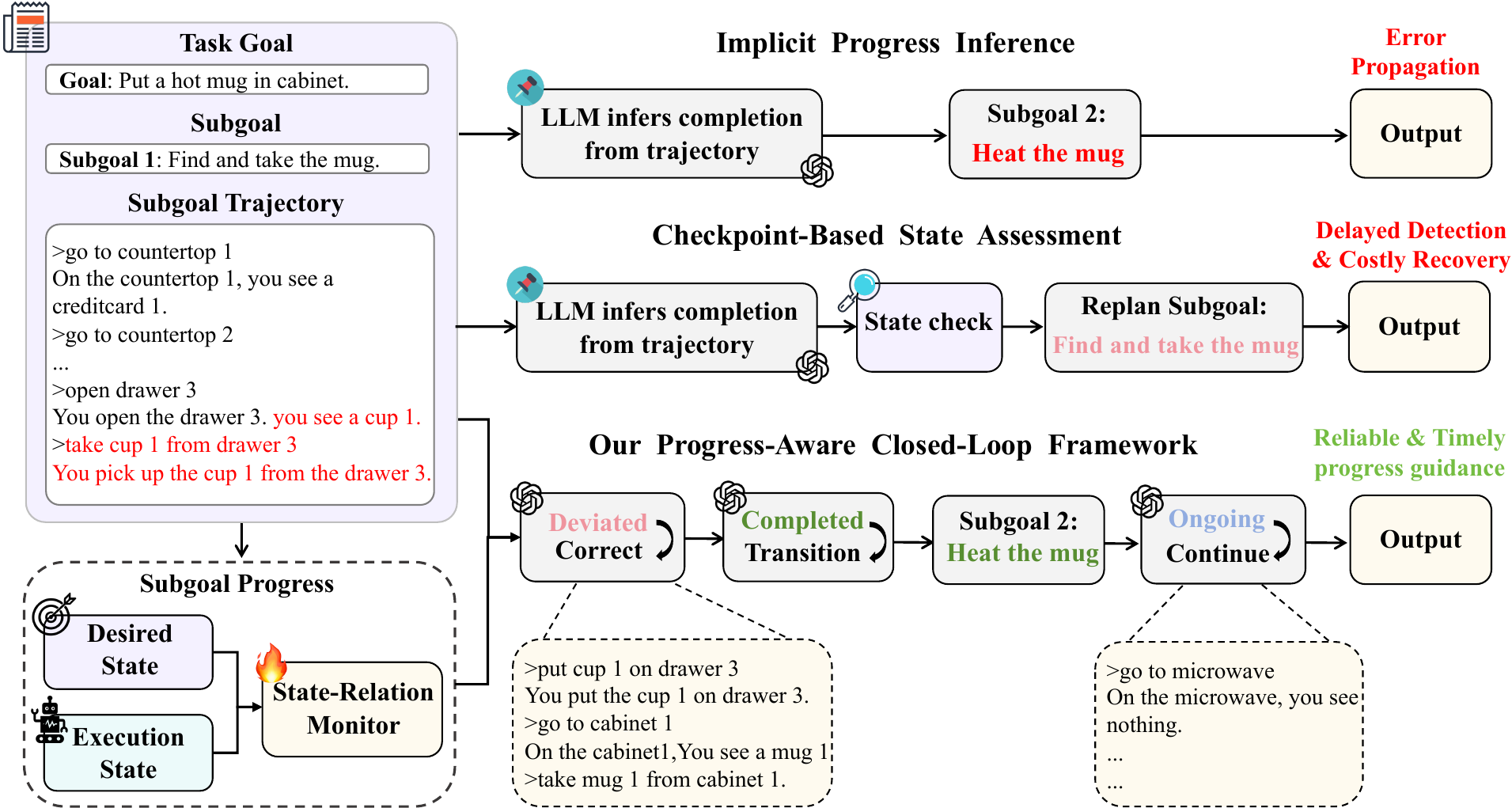}
    \caption{Compared with existing approaches, ProLoop explicitly derives
    subgoal progress from the relation between desired and execution states,
    distinguishes \emph{ongoing}, \emph{deviated}, and \emph{completed} execution,
    and uses the resulting progress to support timely closed-loop adaptation.}
    \label{fig:motivation}
    \vspace{-6pt}
\end{figure}

Building on these observations, recent studies
\cite{sun2023adaplanner,kim-hwang-2025-coex}
address this problem by introducing explicit checks at selected checkpoints to
assess the current execution state and then support replanning
(see Fig.~\ref{fig:motivation}).
These checks form a feedback loop that reduces reliance on implicit progress inference by the LLM.
However, such checkpoint-based feedback cannot timely respond to changes in subgoal progress,
since reassessment occurs only at the next checkpoint.
During robotic execution, this delay can allow erroneous actions
to alter the environment state,
making recovery costly or even infeasible
\cite{kim2025reflact,farag2025conditionalrecovery}.
We therefore ask:
\textbf{How can robots maintain reliable progress awareness throughout subgoal execution
to enable timely closed-loop adaptation?}

A useful intuition comes from how people perform everyday tasks. People assess progress by relating
the current state to the intended outcome before deciding what to do next. Following this intuition,
we define robotic subgoal progress by aggregating structured relations between desired subgoal
conditions and the evolving execution state into \emph{ongoing}, \emph{deviated}, or \emph{completed}.

This formulation raises two key challenges.
\textbf{Constructing comparable desired and execution states.}
A semantic subgoal specifies its desired outcome in natural language, whereas the execution state
must be reconstructed from evolving actions, observations, perception, and contextual information.
Making the two states comparable is difficult,
but essential for accurate subgoal progress assessment.
\textbf{Using progress to guide execution without sacrificing LLM flexibility.}
Even with reliable progress assessment, different progress states call for different responses.
Excessive intervention during ongoing execution may disrupt useful LLM reasoning and action generation,
whereas insufficient intervention after deviation may allow errors to persist.
Moreover, completion should be leveraged beyond stage termination, as its associated outcomes must be selectively transferred to support subsequent planning.

To address these challenges, we propose ProLoop, a
Progress-Aware Closed-Loop framework for long-horizon robotic task planning.
For each active subgoal generated by the \textbf{LLM-Based Subgoal Planner},
\textbf{Structured State Construction} first converts the subgoal's desired outcome
and evolving execution information into comparable desired and execution states
under a shared task-state schema.
A learned \textbf{State-Relation Monitor} then assesses these condition-level relations
and derives the active subgoal progress as ongoing, deviated, or completed.
The resulting progress drives \textbf{Progress-Guided Adaptation}:
ongoing execution preserves the underlying LLM's reasoning process,
deviation triggers targeted correction, and completion permits transition to the next subgoal.
Additionally, upon completion, \textbf{Completion-Gated Memory Update}
uses planner-generated retention cues to selectively summarize task-relevant
execution information for subsequent planning.

To assess the effectiveness of ProLoop, we conduct experiments on the AgentBoard
\cite{zhang2024agentboard} ALFWorld household
planning benchmark, which contains six task categories with easy and hard difficulty levels.
Empirical results demonstrate improved overall performance, with especially substantial gains on
hard tasks. We further validate ProLoop on a physical robot.
Our contributions are threefold:
\begin{enumerate}

    \item We propose \textbf{ProLoop}, a unified progress-aware closed-loop framework for robotic task planning 
    that integrates structured state construction, state-relation monitoring, and progress-guided adaptation.

    \item We define explicit subgoal progress through structured relations between desired and
    execution states, distinguishing \emph{ongoing}, \emph{deviated}, and \emph{completed} progress.

    \item We demonstrate that ProLoop improves task success rate and progress rate on ALFWorld while
    requiring fewer environment interactions. We further validate its feasibility on a physical robot.

\end{enumerate}

%%%%%%%%%%%%%%%%%%%%%%%%%%%%%%%%%%%%%%%%%%%%%%%%%%%%%%%%%%%%%%%%%%%%%%%%%%%%%%%%
\section{Related Work}

\subsection{LLMs for Task Planning}

Recent studies have leveraged the reasoning and long-horizon planning
capabilities of LLMs for robotic tasks \cite{huang2024planning}.
LID \cite{li2022lid} formulates observations, goals, and interaction histories
as sequential representations for policy prediction, while ReAct
\cite{yao2023react} interleaves reasoning and action generation.
Code-based methods, including Code as Policies \cite{liang2023codeaspolicies}
and ProgPrompt \cite{singh2023progprompt}, further represent plans as
executable programs to improve structural consistency and interpretability.
In embodied settings, SayCan \cite{ichter2023saycan} grounds language-based
reasoning in robot affordances, and LLM-Planner \cite{song2023llmplanner}
extends LLM-based planning to embodied decision-making tasks.
These methods generate and ground multi-step plans, but long-horizon
dependencies and growing interaction histories motivate more structured task
organization across successive subgoals.

\subsection{Hierarchical Task Planning}

Hierarchical approaches reduce long-horizon complexity by organizing tasks
around intermediate semantic objectives. HiP \cite{ajay2023hip} coordinates
language planning, visual prediction, and low-level control across multiple
abstraction levels, whereas DELTA \cite{liu2025delta} uses LLM-generated
subgoals to construct smaller problems for an automated planner. STEP Planner
\cite{zhou2025stepplanner} organizes planning through a hierarchical subgoal
tree, while HiAgent \cite{hu2025hiagent} treats semantic subgoals as
working-memory chunks and compresses completed trajectories. Together, these
methods establish subgoals as effective
units for task decomposition and context organization.
However, subgoal progress is still often inferred implicitly by the LLM, making execution prone to error propagation.

\subsection{Feedback-Aware Task Planning}

Feedback-aware methods incorporate execution information to improve planning
robustness. Inner Monologue \cite{huang2023innermonologue} integrates
environment feedback into closed-loop language planning, while AdaPlanner
\cite{sun2023adaplanner} uses predicted feedback and assertions to refine
plans during execution. CoEx \cite{kim-hwang-2025-coex} verifies subgoal
experiences to update an adaptive belief state, and LLaMAR
\cite{nayak2024llamar} combines subtask verification with execution
correction. KnowLoop \cite{zheng2024knowloop} further detects whether
executed subgoals have succeeded using an uncertainty-aware failure detector.
Collectively, these methods support plan refinement,
belief updating, verification, and failure detection.

In summary, prior work has advanced LLM planning through action generation,
hierarchical decomposition, and feedback-based adaptation. However, subgoal
progress is often inferred implicitly or assessed through discrete feedback
mechanisms, leaving continuous progress awareness underexplored.

\section{Preliminaries}

\subsection{Planning Framework}

A task is characterized by a tuple
$\mathcal{P}=\langle G,S,O,A,T\rangle$, where $G$ denotes the goal condition,
$S$ and $O$ represent the state and observation spaces, $A$ is the set of
available actions, and $T:S\times A\rightarrow S$ specifies the state
transition induced by action execution. Planning aims to obtain a policy
that drives the system from an
initial state to a state satisfying $G$. In LLM-based planning, these task
elements are commonly expressed through a natural-language description $D$
provided to the model.

\subsection{Hierarchical Subgoal Execution}

In hierarchical LLM planning, a long-horizon task is executed
through a sequence of semantic subgoals. At stage $k$, the planner generates an active
subgoal $g_k$ as an intermediate objective toward the overall goal $G$.
Executing $g_k$ involves multiple environment interactions, forming a local
trajectory
$\tau_k^t=[(a_{k,1},o_{k,1}),\ldots,(a_{k,t},o_{k,t})]$.
Let $m_i$ denote the information retained from the trajectory of
a previous subgoal $g_i$. The cross-subgoal context before executing $g_k$ is
therefore
$M_k=[(g_1,m_1),\ldots,(g_{k-1},m_{k-1})]$.
Accordingly, the working context of the active subgoal is represented as
$W_k^t=[G,M_k,g_k,\tau_k^t]$.

\begin{figure*}[t!]
    % Shift the float inside the top margin, using space from its lower gap.
    \vspace*{4.5pt}
    \centering
    \includegraphics[width=\textwidth]{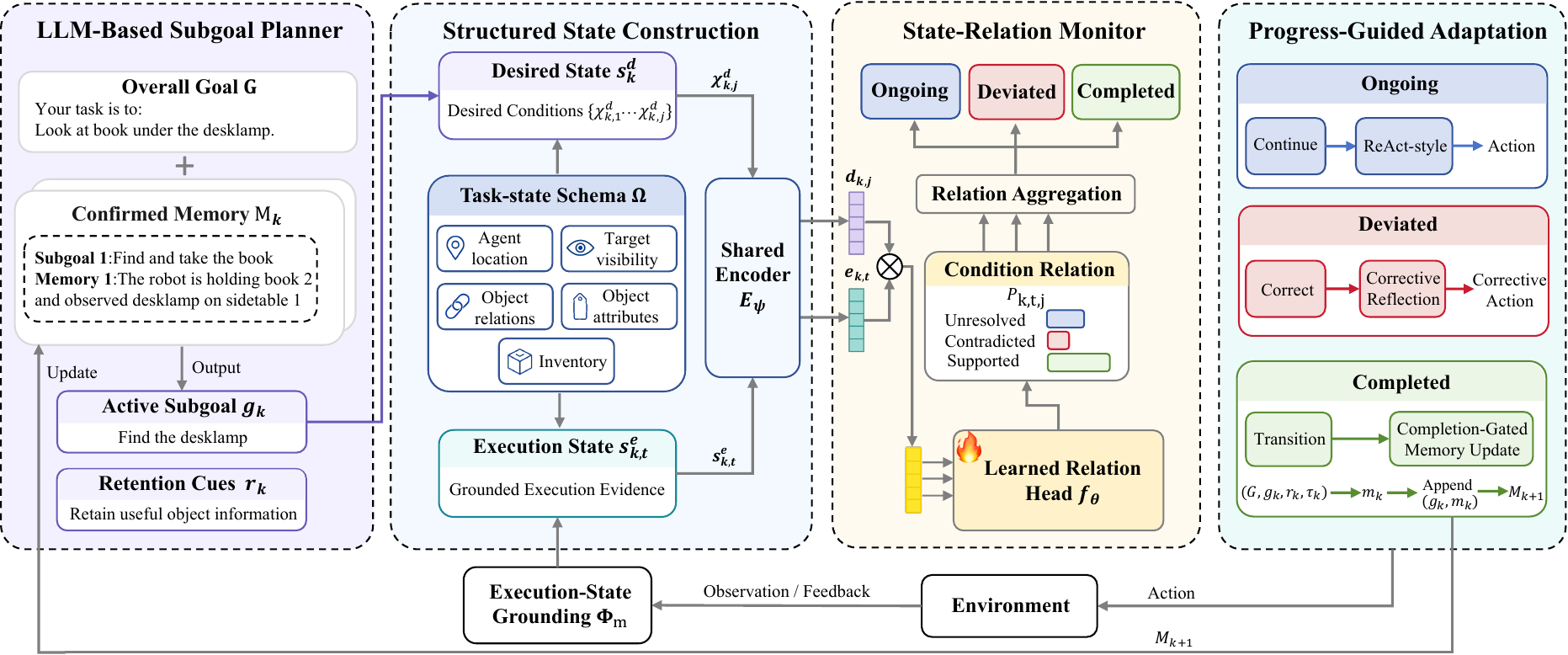}
    \caption{Overview of ProLoop. The LLM-Based Subgoal Planner generates the active subgoal and retention cues. Structured State Construction derives comparable desired and execution states, the State-Relation Monitor infers subgoal progress from their structured relations, and Progress-Guided Adaptation regulates continuation, targeted correction, or subgoal transition, with memory updated upon completion.}
    \label{fig:framework}
    \vspace{-4mm}
    \vspace{-4.5pt}
\end{figure*}

\section{Methodology}
\label{sec:methodology}

In this section, we detail our proposed framework, \textbf{ProLoop}. As illustrated in
Fig.~\ref{fig:framework}, the framework forms a closed loop around each active subgoal
through structured state construction, state-relation monitoring, progress-guided adaptation,
and completion-gated memory update.
The LLM-Based Subgoal Planner generates an active semantic subgoal and retention cues.
Structured State Construction represents the desired subgoal outcome and evolving
execution information under a shared task-state schema.
The State-Relation Monitor compares each desired condition against the current execution
state and classifies the resulting relation as \emph{supported}, \emph{unresolved}, or
\emph{contradicted}. These relations are then aggregated into \emph{ongoing},
\emph{deviated}, or \emph{completed} progress.
Progress-Guided Adaptation maps the resulting progress state to continuation, correction,
or subgoal transition. Upon completion, Completion-Gated Memory Update uses the retention
cues to summarize the local trajectory into a new memory entry before the next subgoal is generated.

\subsection{LLM-Based Subgoal Planner}

At stage $k$, the LLM-based planner generates an active semantic subgoal
$g_k$ and retention cues $r_k$:
\begin{equation}
    (g_k,r_k)
    =
    P_{\mathrm{LLM}}(G,M_k,I_P),
    \label{eq:subgoal_planner}
\end{equation}
where $G$ and $M_k$ are defined in Sec.~III, and $I_P$ contains the
planning instruction, in-context demonstrations, and output format. The
subgoal $g_k$ specifies the current intermediate objective, whereas $r_k$
indicates information that may be useful to subsequent subgoals. ProLoop
generates one active subgoal at a time, allowing later planning decisions to
condition on confirmed outcomes from earlier stages.

\subsection{Structured State Construction}

Structured State Construction converts the desired subgoal outcome and evolving
execution information into comparable desired and execution states under a shared
task-state schema $\Omega$. It produces a desired state $s_k^d$ and an execution state
$s_{k,t}^e$ for subsequent relation assessment by the monitor.

\textbf{Desired-State Construction.}
The schema $\Omega$ contains general task-state field types, including agent
location, target visibility, object attributes, object
relations and inventory. Given the active subgoal, an LLM-based constructor
generates the following desired state:
\begin{equation}
    s_k^d
    =
    C_{\mathrm{LLM}}(g_k,\Omega,I_C),
    \label{eq:desired_state}
\end{equation}
where $I_C$ specifies the construction instruction. The resulting state is a
sparse set of $J_k$ desired conditions,
$s_k^d=\{\chi_{k,j}^d\}_{j=1}^{J_k}$. Each condition is represented by the following tuple:
\begin{equation}
    \chi_{k,j}^d
    =
    \langle f_{k,j},b_{k,j},v_{k,j}^d\rangle .
    \label{eq:desired_condition}
\end{equation}
The three tuple elements denote, respectively, the field type, entity binding,
and desired value. Desired conditions may retain category-level bindings,
whereas execution states record grounded instances (e.g., \texttt{mug} versus
\texttt{mug 1}).

\begin{figure*}[t!]
    \vspace*{4.5pt}
    \centering
    \includegraphics[width=0.85\textwidth]{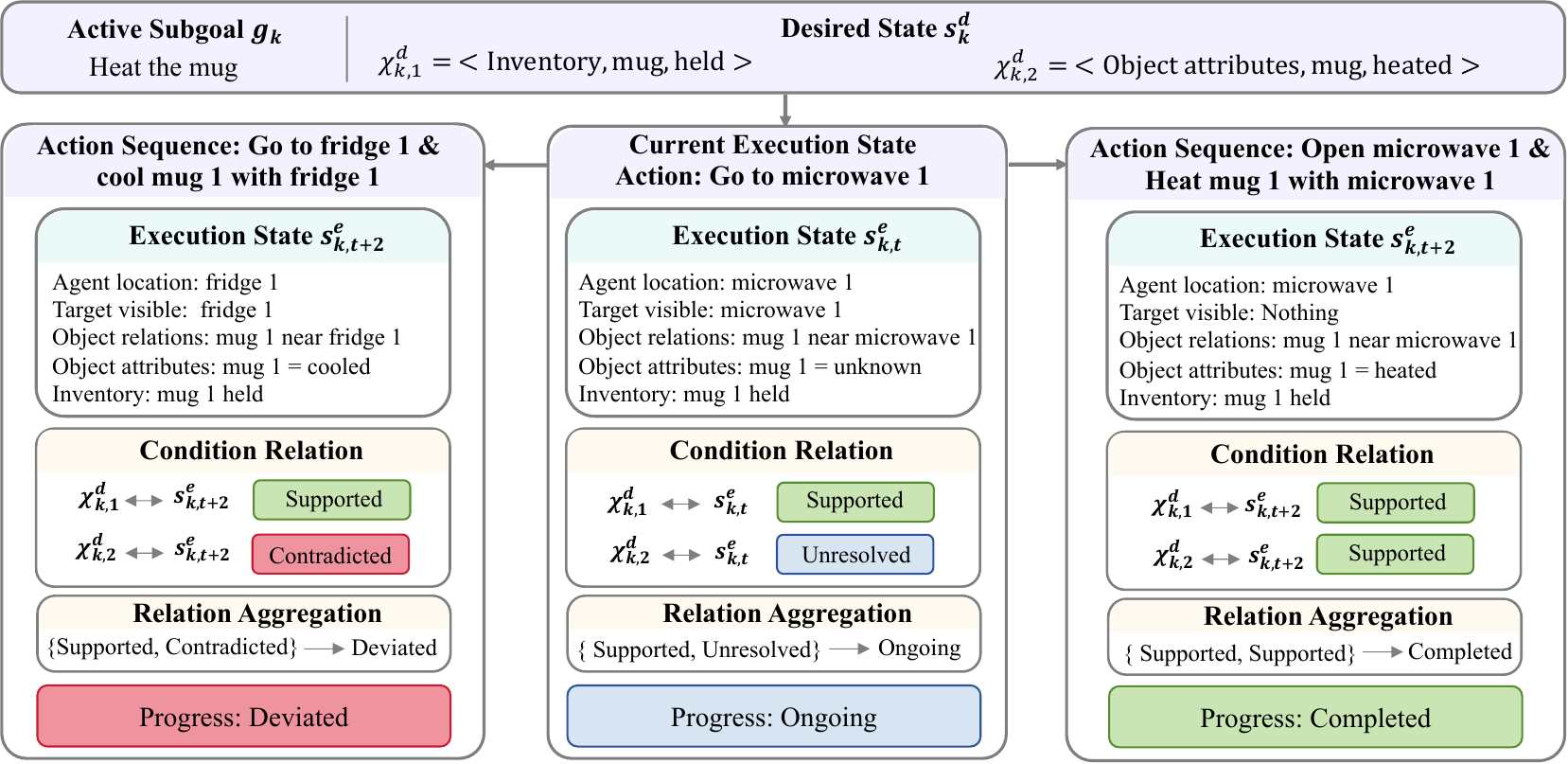}
    \caption{Worked example of state-relation monitoring and progress derivation.
    For the active subgoal \emph{Heat the mug}, desired conditions are compared
    with the corresponding execution state, and the resulting relations are
    aggregated to derive \emph{ongoing}, \emph{deviated}, or \emph{completed}
    subgoal progress.}
    \label{fig:proloop3}
    \vspace{-12pt}
\end{figure*}

\textbf{Execution-State Grounding.}
Let $o_{k,t}$ denote the observation and available execution feedback after
action $a_{k,t}$. A modality-specific grounding function $\Phi_m$ extracts
task-relevant fields:
\begin{equation}
    \Delta s_{k,t}^e
    =
    \Phi_m(a_{k,t}, o_{k,t}, g_k, \Omega).
    \label{eq:evidence_grounding}
\end{equation}

The active subgoal guides task-relevant grounding, while $\Omega$ specifies
the output schema. At task start, $s_{1,0}^e=\varnothing$; for $k>1$,
$s_{k,0}^e=s_{k-1,T_{k-1}}^e$, where $T_{k-1}$ is the final step of
$g_{k-1}$. The execution state is updated as follows:
\begin{equation}
    s_{k,t}^e
    =
    \mathcal{U}(s_{k,t-1}^e,\Delta s_{k,t}^e),
    \label{eq:execution_update}
\end{equation}
where $\mathcal{U}$ incorporates new information and removes invalidated facts.
For constrained text environments, $\Phi_m$ uses deterministic parsing over
agent-visible observations and action feedback. For physical execution, visual
observations are grounded by a VLM with robot and skill feedback. Persistent
facts are retained unless invalidated, while transient fields are refreshed
from current observations and unavailable values are marked unknown.
Let $\operatorname{Inv}(o_{k,t})$ indicate an invalid action. Such feedback
directly yields \emph{deviated} progress.

\subsection{State-Relation Monitor}

As illustrated in Fig.~\ref{fig:proloop3}, the monitor compares each desired
condition with the current execution state and aggregates the resulting
relations into subgoal progress.
A \emph{supported} relation indicates that available evidence establishes
the desired condition.
An \emph{unresolved} relation indicates that the condition is not yet
established and no conflicting evidence is available.
A \emph{contradicted} relation indicates explicit evidence of a task-relevant
conflict with the desired condition.
For instance, when the
desired attribute is heated, an unknown thermal-state value is unresolved,
whereas an explicitly recorded cooled value is contradicted.

The execution state $s_{k,t}^e$ is serialized in a fixed schema order and
encoded once per interaction, while each desired condition $\chi_{k,j}^d$
is encoded individually using the same pretrained semantic encoder:
\begin{equation}
    \mathbf{d}_{k,j}
    =
    E_\psi\!\left(\operatorname{Ser}(\chi_{k,j}^d)\right),
    \qquad
    \mathbf{e}_{k,t}
    =
    E_\psi\!\left(\operatorname{Ser}(s_{k,t}^e)\right),
    \label{eq:state_embeddings}
\end{equation}
where $\operatorname{Ser}(\cdot)$ converts a structured input into a canonical
textual sequence: execution-state fields follow the fixed schema order, while
each desired condition preserves its field type, entity binding, and target
value. For each desired condition, we construct a pairwise relation
representation using the two embeddings, their absolute difference, and their
element-wise interaction to capture condition--state compatibility:
\begingroup
\setlength{\abovedisplayskip}{4pt}
\begin{equation}
    \mathbf x_{k,t,j}
    =
    [
    \mathbf d_{k,j};
    \mathbf e_{k,t};
    |\mathbf d_{k,j}-\mathbf e_{k,t}|;
    \mathbf d_{k,j}\odot\mathbf e_{k,t}
    ].
    \label{eq:relation_feature}
\end{equation}
\endgroup
Here, $[\cdot\,;\cdot]$ denotes concatenation and $\odot$ denotes the
element-wise product. A shared lightweight MLP relation head $f_\theta$
is trained to assess condition--state compatibility using these pairwise
features, with category-to-instance correspondence learned implicitly through
supervision from the three relation labels. The head predicts the relation
probabilities:
\begin{equation}
    \mathbf p_{k,t,j}
    =
    \operatorname{softmax}
    \bigl(f_\theta(\mathbf x_{k,t,j})\bigr).
    \label{eq:relation_prediction}
\end{equation}
The probability vector contains the following components:
\begin{equation}
    \mathbf p_{k,t,j}
    =
    [
    p_{k,t,j}^{\mathrm{sup}},
    p_{k,t,j}^{\mathrm{unr}},
    p_{k,t,j}^{\mathrm{con}}
    ].
    \label{eq:relation_distribution}
\end{equation}
The entries correspond to the predicted probabilities of \emph{supported},
\emph{unresolved}, and \emph{contradicted}, respectively.
The predicted condition relation is determined as follows:
\begin{equation}
    \ell_{k,t,j}
    =
    \arg\max_{c\in\mathcal Q} p_{k,t,j}^{c},
    \qquad
    \mathcal Q
    =
    \{\mathrm{sup},\mathrm{unr},\mathrm{con}\}.
    \label{eq:condition_relation}
\end{equation}
We derive subgoal progress by aggregating the predicted condition relations
and checking for invalid-action feedback:
\begin{equation}
    \rho_{k,t}
    =
    \begin{cases}
        \textit{deviated},
        & \operatorname{Inv}(o_{k,t})\ \lor\ \exists j:\ \ell_{k,t,j}=\mathrm{con},\\
        \textit{completed},
        & \neg\operatorname{Inv}(o_{k,t})\ \land\ \forall j:\ \ell_{k,t,j}=\mathrm{sup},\\
        \textit{ongoing},
        & \text{otherwise}.
    \end{cases}
    \label{eq:progress_state}
\end{equation}
We adopt a conservative aggregation rule: contradictions trigger correction,
while transition requires all desired conditions to be supported.

\begin{figure*}[t!]
    \vspace*{4.5pt}
    \centering
    \includegraphics[width=\textwidth]{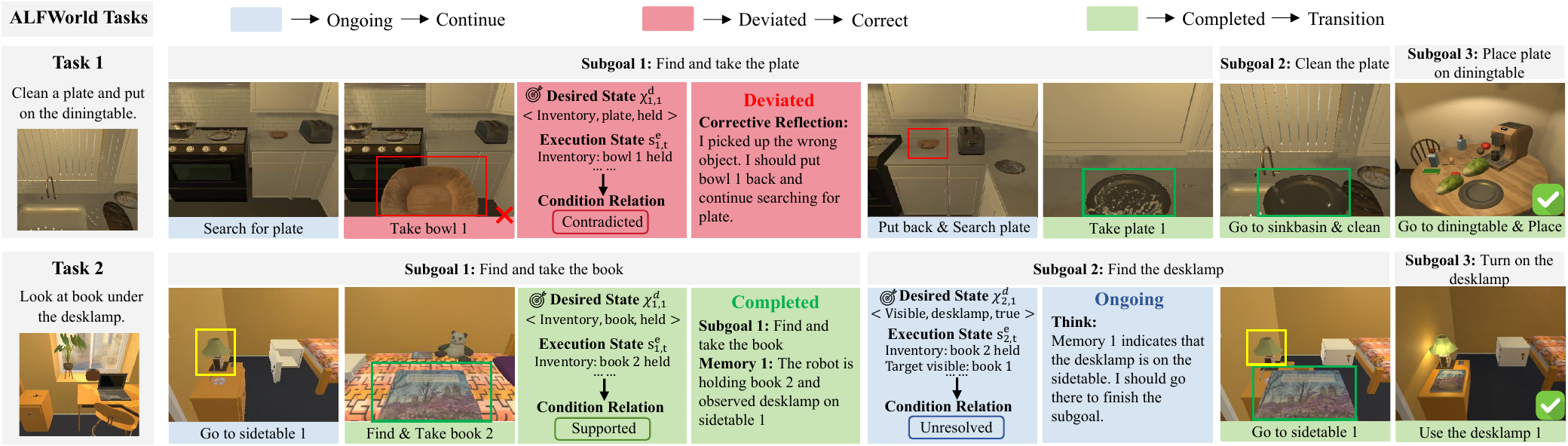}
    \caption{Representative ALFWorld executions with ProLoop.
    Ongoing progress allows execution to continue without additional intervention,
    deviated progress triggers targeted correction, and completed progress enables
    subgoal transition and memory update.}
    \label{fig:case_study}
    \vspace{-12pt}
\end{figure*}

\subsection{Progress-Guided Adaptation}

The derived progress state regulates the underlying executor by selecting
one of three control actions:
\begingroup
\setlength{\abovedisplayskip}{4pt}
\begin{equation}
    \mathcal A(\rho_{k,t})
    =
    \begin{cases}
        \textsc{Continue},
        & \rho_{k,t}=\textit{ongoing},\\
        \textsc{Correct},
        & \rho_{k,t}=\textit{deviated},\\
        \textsc{Transition},
        & \rho_{k,t}=\textit{completed}.
    \end{cases}
    \label{eq:adaptation}
\end{equation}
\endgroup

\textbf{Continue.}
For an ongoing subgoal, ProLoop preserves the underlying ReAct-style
think--act loop without introducing an additional intervention prompt. The
executor continues reasoning from the active subgoal, cross-subgoal memory,
and local trajectory, with each interaction re-evaluated by the State-Relation Monitor.

\textbf{Correct.}
For a deviated subgoal, ProLoop constructs a correction prompt from the
detected conflicting execution information and available failure feedback.
The prompt is injected into the executor's working context to elicit a
\emph{Corrective Reflection}, which is retained while the subgoal remains
deviated and guides the generation of a \emph{Corrective Action} without
changing the active subgoal. Each resulting interaction is re-evaluated by
the State-Relation Monitor, and the corrective reflection is removed once
the progress leaves the \emph{deviated} state.

\textbf{Transition.}
For a completed subgoal, ProLoop terminates the current execution stage and
invokes the Completion-Gated Memory Update in Sec.~IV-E to summarize the
completed local trajectory into a memory entry. If the overall goal remains
unsatisfied, control returns to the planner to generate $g_{k+1}$; otherwise,
execution terminates.

\subsection{Completion-Gated Memory Update}

Cross-subgoal memory is updated only after the derived progress state indicates
completion. An LLM summarizes the complete local trajectory $\tau_k$ of $g_k$,
guided by the retention cues $r_k$ to retain task-relevant information:
\begingroup
\setlength{\abovedisplayskip}{4pt}
\begin{equation}
    m_k
    =
    S_{\mathrm{LLM}}
    (G,g_k,r_k,\tau_k,I_M),
    \label{eq:memory_summary}
\end{equation}
\endgroup
where $I_M$ specifies the memory-update instruction. The retention cues guide
the summary but do not prevent newly discovered task-relevant information from
being retained.

The new subgoal--memory pair is appended to the existing context
to support subsequent subgoal planning:
\begin{equation}
    M_{k+1}
    =
    M_k\oplus(g_k,m_k).
    \label{eq:memory_update}
\end{equation}
Here, $\oplus$ appends a new entry without merging previous memories. Only the
cue-guided summary is carried forward; the full local trajectory is not
appended, reducing cross-subgoal context growth.

\subsection{Offline Relation Learning}

The State-Relation Monitor is trained offline using trajectories collected
only from training environments; no OOD evaluation instances are used for
relation-head training. Human-written annotation guidelines define the
semantics of the three condition relations. Guided by these guidelines,
Qwen3.8-Max serves as the offline annotator and assigns each condition--state
pair $(\chi_{k,j}^d,s_{k,t}^e)$ a relation label in
$\mathcal Q=\{\mathrm{sup},\mathrm{unr},\mathrm{con}\}$. A randomly sampled
subset of the generated annotations is manually verified for consistency
with the same guidelines. The resulting labels provide condition-level
supervision for the relation head; the LLM annotator is used only offline and
is not involved during execution.

Given the predicted relation distribution $\mathbf p_{k,t,j}$, the relation
head is optimized using weighted cross-entropy:
\begin{equation}
    \mathcal L_{\mathrm{rel}}
    =
    -
    \sum_{k,t,j}
    \sum_{c\in\mathcal Q}
    w_c\,y_{k,t,j}^{c}\log p_{k,t,j}^{c},
    \label{eq:relation_training}
\end{equation}
where $y_{k,t,j}^{c}$ is the annotated target and $w_c$ compensates for class
imbalance. We freeze the pretrained semantic encoder and train only the MLP
relation head.

\section{Experiments}
\label{sec:experiments}

In this section, we evaluate ProLoop on the AgentBoard ALFWorld benchmark under different task difficulties and LLM backbones. The results demonstrate consistent improvements in overall task performance, with particularly pronounced gains on Hard tasks involving more subgoals.

\begin{table*}[!tb]
% Keep the caption inside the top margin without increasing the float height.
\vspace*{5pt}
\caption{Performance of ProLoop and baselines on the AgentBoard ALFWorld OOD split. Results for both backbones are reported as mean $\pm$ standard deviation over three runs with different random seeds.}
\label{tab:main_results}
\centering
\small
\setlength{\tabcolsep}{6.3pt}
\begin{tabular}{l|ccc|cc|cc}
\toprule
\multirow[c]{2}{*}[-2.5pt]{\textbf{Method}} &
\multicolumn{3}{c}{\textbf{Overall (\%)}} &
\multicolumn{2}{c}{\textbf{Easy (\%)}} &
\multicolumn{2}{c}{\textbf{Hard (\%)}} \\
\cmidrule(lr){2-4}\cmidrule(lr){5-6}\cmidrule(lr){7-8}
& \textbf{SR} \(\uparrow\) & \textbf{PR} \(\uparrow\) & \textbf{GA} \(\uparrow\) &
\textbf{SR} \(\uparrow\) & \textbf{PR} \(\uparrow\) &
\textbf{SR} \(\uparrow\) & \textbf{PR} \(\uparrow\) \\
\midrule
\multicolumn{8}{c}{\emph{Model: Qwen3-14B}}\\
\midrule
Plan-Only
& \(57.71{\scriptstyle\pm2.46}\) & \(67.77{\scriptstyle\pm1.42}\) & \(82.02{\scriptstyle\pm1.29}\)
& \(81.94{\scriptstyle\pm1.96}\) & \(84.26{\scriptstyle\pm1.73}\) & \(52.42{\scriptstyle\pm3.09}\) & \(64.17{\scriptstyle\pm1.98}\) \\
Reason-Only
& \(83.58{\scriptstyle\pm1.06}\) & \(91.21{\scriptstyle\pm0.44}\) & \(93.18{\scriptstyle\pm0.38}\)
& \(94.44{\scriptstyle\pm5.20}\) & \(95.37{\scriptstyle\pm4.72}\) & \(81.21{\scriptstyle\pm1.87}\) & \(90.30{\scriptstyle\pm0.95}\) \\
Reason-Hierarchical
& \(69.90{\scriptstyle\pm1.41}\) & \(79.71{\scriptstyle\pm1.41}\) & \(94.53{\scriptstyle\pm0.06}\)
& \(90.28{\scriptstyle\pm1.96}\) & \(91.20{\scriptstyle\pm0.65}\) & \(65.45{\scriptstyle\pm1.96}\) & \(77.20{\scriptstyle\pm1.75}\) \\
Reason-Replan
& \(75.62{\scriptstyle\pm2.54}\) & \(83.46{\scriptstyle\pm1.56}\) & \(91.88{\scriptstyle\pm1.08}\)
& \cellcolor{gray!15}\(\mathbf{100.00}{\scriptstyle\pm0.00}\) & \cellcolor{gray!15}\(\mathbf{100.00}{\scriptstyle\pm0.00}\) & \(70.30{\scriptstyle\pm3.09}\) & \(79.85{\scriptstyle\pm1.90}\) \\
\textbf{Ours}
& \cellcolor{gray!15}\(\mathbf{96.52{\scriptstyle\pm0.70}}\) & \cellcolor{gray!15}\(\mathbf{97.39{\scriptstyle\pm0.18}}\) & \cellcolor{gray!15}\(\mathbf{98.23{\scriptstyle\pm0.05}}\)
& \(95.83{\scriptstyle\pm0.00}\) & \(98.61{\scriptstyle\pm0.00}\) & \cellcolor{gray!15}\(\mathbf{96.67{\scriptstyle\pm0.86}}\) & \cellcolor{gray!15}\(\mathbf{97.12{\scriptstyle\pm0.21}}\) \\
\midrule
\multicolumn{8}{c}{\emph{Model: GPT-5.6}}\\
\midrule
Plan-Only
& \(84.33{\scriptstyle\pm3.39}\) & \(90.90{\scriptstyle\pm2.08}\) & \(94.54{\scriptstyle\pm0.25}\)
& \(93.06{\scriptstyle\pm5.20}\) & \(93.98{\scriptstyle\pm4.29}\) & \(82.42{\scriptstyle\pm3.09}\) & \(90.23{\scriptstyle\pm1.70}\) \\
Reason-Only
& \(87.06{\scriptstyle\pm1.76}\) & \(92.29{\scriptstyle\pm0.58}\) & \(96.87{\scriptstyle\pm0.26}\)
& \(93.06{\scriptstyle\pm1.96}\) & \(93.06{\scriptstyle\pm1.96}\) & \(85.76{\scriptstyle\pm2.39}\) & \(92.12{\scriptstyle\pm0.95}\) \\
Reason-Hierarchical
& \(89.30{\scriptstyle\pm1.53}\) & \(92.54{\scriptstyle\pm1.19}\) & \(97.61{\scriptstyle\pm0.33}\)
& \(90.28{\scriptstyle\pm1.96}\) & \(91.67{\scriptstyle\pm0.00}\) & \(89.09{\scriptstyle\pm1.48}\) & \(92.73{\scriptstyle\pm1.45}\) \\
Reason-Replan
& \(92.54{\scriptstyle\pm1.06}\) & \(95.52{\scriptstyle\pm0.61}\) & \(97.34{\scriptstyle\pm0.58}\)
& \cellcolor{gray!15}\(\mathbf{100.00}{\scriptstyle\pm0.00}\) & \cellcolor{gray!15}\(\mathbf{100.00}{\scriptstyle\pm0.00}\) & \(90.91{\scriptstyle\pm1.29}\) & \(94.55{\scriptstyle\pm0.74}\) \\
\textbf{Ours}
& \cellcolor{gray!15}\(\mathbf{97.51{\scriptstyle\pm0.35}}\) & \cellcolor{gray!15}\(\mathbf{97.64{\scriptstyle\pm0.18}}\) & \cellcolor{gray!15}\(\mathbf{99.01{\scriptstyle\pm0.28}}\)
& \cellcolor{gray!15}\(\mathbf{100.00{\scriptstyle\pm0.00}}\) & \cellcolor{gray!15}\(\mathbf{100.00{\scriptstyle\pm0.00}}\) & \cellcolor{gray!15}\(\mathbf{96.97{\scriptstyle\pm0.43}}\) & \cellcolor{gray!15}\(\mathbf{97.12{\scriptstyle\pm0.21}}\) \\
\bottomrule
\end{tabular}
\vspace{-5pt}
\end{table*}

\begin{table}[!tb]
\caption{Environment interaction and token efficiency on the AgentBoard ALFWorld OOD split with Qwen3-14B. Results are reported as mean $\pm$ standard deviation over three runs.}
\label{tab:efficiency_results}
\centering
\small
\setlength{\tabcolsep}{4.5pt}
\begin{tabular}{lcc}
\toprule
\textbf{Method} &
\textbf{Env. steps} \(\downarrow\) &
\textbf{Tokens / episode} \(\downarrow\) \\
\midrule
Plan-Only & \(20.61{\scriptstyle\pm0.47}\) & \(31,676{\scriptstyle\pm875}\) \\
Reason-Only & \(14.98{\scriptstyle\pm0.23}\) & \(54,713{\scriptstyle\pm2,629}\) \\
Reason-Hierarchical & \(17.25{\scriptstyle\pm0.10}\) & \(35,950{\scriptstyle\pm25}\) \\
Reason-Replan & \(17.29{\scriptstyle\pm0.11}\) & \cellcolor{gray!15}\(\mathbf{8,422{\scriptstyle\pm104}}\) \\
\textbf{Ours} & \cellcolor{gray!15}\(\mathbf{10.14{\scriptstyle\pm0.13}}\) & \(25,191{\scriptstyle\pm171}\) \\
\bottomrule
\end{tabular}\par
\vspace{-12pt}
\end{table}

\subsection{Experimental Setup}

\textbf{Environment and Dataset.}
We conduct our evaluation on the out-of-distribution (OOD) split of ALFWorld
provided by AgentBoard~\cite{zhang2024agentboard,shridhar2021alfworld}.
The benchmark contains six household task categories:
\textit{Pick \& Place}, \textit{Examine in Light}, \textit{Clean \& Place},
\textit{Heat \& Place}, \textit{Cool \& Place}, and
\textit{Pick Two \& Place}, comprising 134 unseen instances in total.
Following AgentBoard's difficulty labels, the evaluation set contains
24 Easy and 110 Hard tasks, with three and four benchmark-defined subgoals,
respectively. Each episode is capped at 30 environment interactions, and
all methods follow the same data split, interaction budget, and evaluation
protocol for a consistent comparison across all methods.

\textbf{Compared Methods.}
We compare ProLoop with four representative paradigms for LLM-based long-horizon planning:
\textbf{1) Plan-Only:} AgentBoard's VanillaAgent directly generates environment actions without explicit reasoning.
\textbf{2) Reason-Only:} ReAct~\cite{yao2023react}, which interleaves step-wise reasoning and action generation.
\textbf{3) Reason-Hierarchical:} HiAgent~\cite{hu2025hiagent}, which organizes long interaction histories through semantic subgoals and hierarchical working memory.
\textbf{4) Reason-Replan:} AdaPlanner~\cite{sun2023adaplanner}, which adapts and refines plans using closed-loop execution feedback.
We evaluate the methods using Qwen3-14B and GPT-5.6 as LLM backbones. All ALFWorld experiments are conducted over three runs with different random seeds for each method.

\textbf{Metrics.}
We evaluate all methods from three complementary perspectives:

\begin{itemize}
    \setlength{\itemsep}{1pt}
    \setlength{\parskip}{0pt}

    \item \textbf{Overall Task Performance.}
    Success Rate (SR) measures final task completion, Progress Rate (PR)
    measures the maximum fraction of benchmark-defined progress steps satisfied
    during execution, and Grounding Accuracy (GA) measures the proportion
    of generated actions that are valid under the environment state at the time of execution.

    \item \textbf{Performance by Task Difficulty.}
    We report SR and PR separately on the Easy and Hard subsets.

    \item \textbf{Efficiency.}
    We report the per-episode averages of environment interactions and total
    tokens across all online LLM calls during execution.

\end{itemize}

\subsection{Main Results}

Tables~\ref{tab:main_results} and~\ref{tab:efficiency_results}
report task performance and efficiency, respectively, while
Fig.~\ref{fig:case_study} shows representative ProLoop executions in ALFWorld.
\textbf{1) Overall Task Performance.}
Compared with the strongest corresponding baselines, ProLoop improves overall
SR by 12.94\% and 4.97\%, and PR by 6.18\% and 2.12\% under Qwen3-14B
and GPT-5.6, respectively. It also achieves the highest GA under both backbones,
reaching 98.23\% and 99.01\%. These results show that explicit progress
assessment and adaptation improve reliable long-horizon execution.
Notably, ProLoop with Qwen3-14B outperforms all four GPT-5.6 baseline
methods in overall performance, showing that effective progress-aware
execution can provide substantial gains beyond those obtained from a stronger
backbone alone.
\textbf{2) Performance by Task Difficulty.}
On the 24 Easy tasks, ProLoop with Qwen3-14B achieves 95.83\% SR and
98.61\% PR, slightly below Reason-Replan at 100\%, while GPT-5.6 ProLoop
matches the best result at 100\% on both metrics. The advantage becomes more
pronounced on the 110 Hard tasks, where ProLoop improves SR by 15.46\% and
6.06\%, and PR by 6.82\% and 2.57\% under the two backbones, respectively.
Since Hard tasks require more subgoals and transition decisions, the larger gains
suggest that explicit progress regulation is particularly beneficial when
coordinating more intermediate objectives.
\textbf{3) Efficiency.}
ProLoop requires only 10.14 environment interactions per episode, the lowest
among all methods. Compared with Reason-Only, it reduces environment
interactions by 32.31\% and token usage by 53.96\%. Its 25,191 tokens per
episode are the second lowest overall, indicating a favorable balance between
task performance and resource efficiency. 
\FloatBarrier

\raggedbottom
\section{Analysis and Discussion}
\afterpage{\vspace*{-38pt}}
\begin{figure*}[t!]
    % Keep the figure inside the top margin and reclaim its lower whitespace.
    \vspace*{4.5pt}
    \centering
    \includegraphics[width=\textwidth]{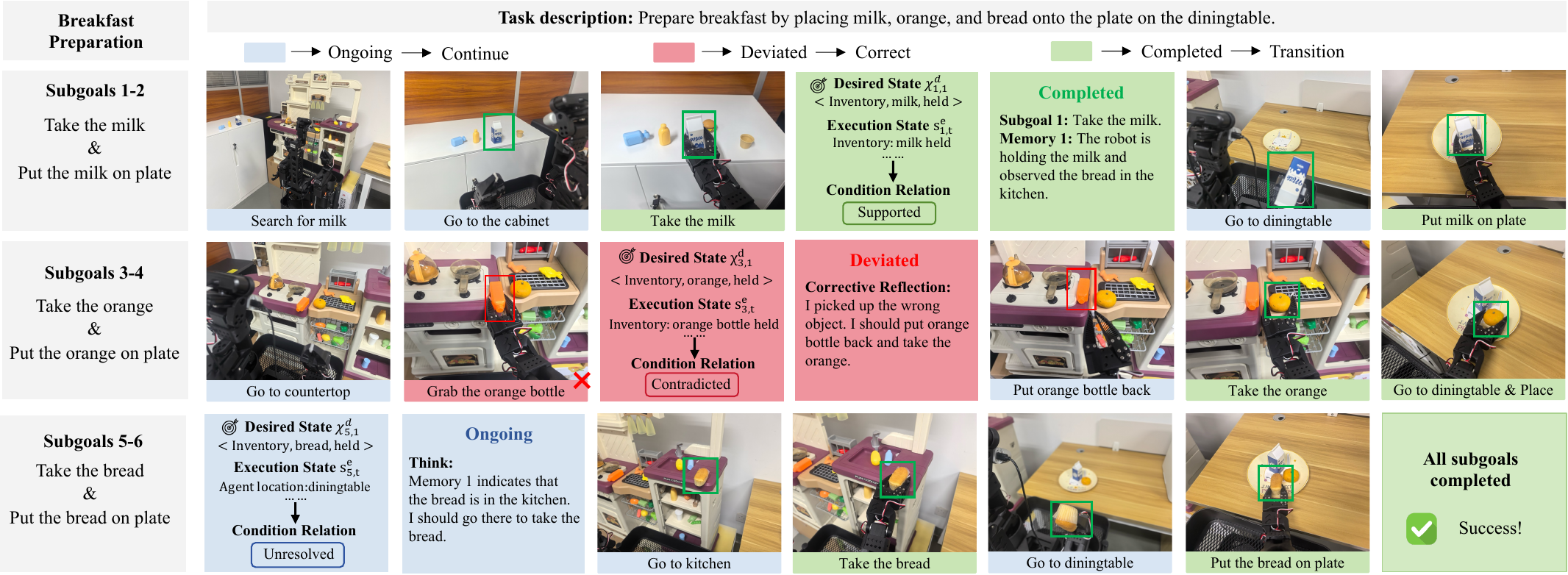}
    \caption{Real-world execution of a breakfast preparation task with ProLoop
    on the dual-arm mobile robot.}
    \label{fig:real_world}
    \vspace{-10.5pt}
\end{figure*}
% Align the table titles; compensate left-column text for the shorter table.
\begin{table*}[t!]
\vspace*{12pt}
\centering
\begin{minipage}[t]{\columnwidth}
\vspace{0pt}
\caption{Subgoal transition reliability under explicit and implicit progress assessment with Qwen3-14B. All values are percentages computed from pooled counts across three runs with different random seeds.}
\label{tab:subgoal_transition_alignment}
\centering
\footnotesize
\renewcommand{\arraystretch}{1.1}
\setlength{\tabcolsep}{2pt}
\begin{tabular}{lcccc}
\toprule
\textbf{Method} &
\textbf{On-time}~\(\uparrow\) &
\textbf{Delayed}~\(\downarrow\) &
\textbf{Premature}~\(\downarrow\) &
\textbf{Missed}~\(\downarrow\) \\
\midrule
Implicit LLM
& 68.70 & 11.94 & 16.08 & 3.29 \\
\rowcolor{gray!15}\cellcolor{white}\textbf{Ours}
& \textbf{98.66} & \textbf{1.34} & \textbf{0.00} & \textbf{0.00}
\\
\bottomrule
\end{tabular}\par
\vspace{0pt}

\end{minipage}\hfill
\begin{minipage}[t]{\columnwidth}
\vspace{0pt}
\caption{Ablation study of ProLoop components on the AgentBoard ALFWorld OOD split with Qwen3-14B. Results are averaged over three runs.}
\label{tab:ablation_results}
\centering
\footnotesize
\renewcommand{\arraystretch}{1.1}
\setlength{\tabcolsep}{3.6pt}
\begin{tabular}{l >{\columncolor{white}[\tabcolsep][8.1pt]}c @{\hspace{16.2pt}} >{\columncolor{white}[8.1pt][\tabcolsep]}c @{\hspace{11.7pt}} c c}
\toprule
\multirow[c]{2}{*}[-2.5pt]{\textbf{Method}} &
\multicolumn{2}{c}{\textbf{Overall (\%)}} &
\textbf{Easy (\%)} &
\textbf{Hard (\%)} \\
\cmidrule(lr){2-3}\cmidrule(lr){4-4}\cmidrule(lr){5-5}
& \textbf{SR} \(\uparrow\) & \textbf{PR} \(\uparrow\) &
\textbf{SR} \(\uparrow\) & \textbf{SR} \(\uparrow\) \\
\midrule
\textbf{Ours full}
& \cellcolor{gray!15}\textbf{96.52} & \cellcolor{gray!15}\textbf{97.39} & 95.83 & \cellcolor{gray!15}\textbf{96.67} \\
w/o Progress Loop
& 85.07 & 91.92 & 95.83 & 82.73 \\
w/o Summary
& 92.54 & 96.08 & \cellcolor{gray!15}\textbf{100.00} & 90.91 \\
w/o Both
& 83.58 & 93.97 & 95.83 & 80.91 \\
w/o Relation Head
& 86.57 & 93.97 & 95.83 & 84.55 \\
\bottomrule
\end{tabular}\par
\vspace{0pt}

\end{minipage}\par
\vspace{-12pt}
\end{table*}

\subsection{Subgoal Transition Reliability}

We compare ProLoop with the
\textit{Implicit LLM} variant (\textit{w/o Progress Loop} in Table~IV).
This variant retains the ReAct-style executor and cross-subgoal memory but
infers completion from the accumulated trajectory and outputs
\textbf{Completed} to trigger transition.
Across three seeds, subgoals are retrospectively labeled by humans with LLM assistance,
excluding those with neither completion nor transition before episode termination.
Transitions are \emph{on-time}, \emph{delayed}, or \emph{premature} if triggered
immediately upon completion, at a later step, or before completion, respectively.
A transition is \emph{missed} if no transition occurs before episode termination
despite subgoal completion.

ProLoop achieves 98.66\% on-time transitions versus 68.70\% for the
Implicit LLM (Table~III). Only 1.34\% are delayed, with no premature or
missed transitions observed in the evaluated runs.
These results suggest that explicit progress assessment supports continued
execution until completion and timely transitions thereafter, enabling
reliable closed-loop adaptation.

\subsection{Ablation Study}

We evaluate component contributions and complementary effects through ablations
(Table~IV).
In \textbf{w/o Progress Loop}, Structured State Construction, State-Relation
Monitoring, and Progress-Guided Adaptation are removed. Overall and Hard-task
SR decrease by 11.45\% and 13.94\%, respectively, while Easy-task SR remains
unchanged, suggesting that explicit progress regulation is more important when
coordinating more subgoals and transitions.
In \textbf{w/o Summary}, completion-gated summaries are replaced with the full
local trajectory $\tau_k$. Overall and Hard-task SR drop by 3.98\% and 5.76\%,
despite a slight Easy-task gain, suggesting that selective summarization becomes
more valuable as cross-subgoal context accumulates.
The \textbf{w/o Both} variant combines these changes, further reducing overall
and Hard-task SR to 83.58\% and 80.91\%, confirming that progress regulation
and completion-gated memory provide complementary benefits.
Finally, \textbf{w/o Relation Head} replaces the learned relation head with
direct LLM judgment under the same desired conditions and structured execution
states. Overall and Hard-task SR decrease by 9.95\% and 12.12\%, respectively,
supporting the benefit of learned relation assessment beyond structured state
representation alone.

\subsection{Real-World Robotic Experiment}

We further validate ProLoop on a dual-arm mobile robot equipped with a RealSense D435i
head camera, using Qwen3.8-flash as the VLM backbone. The robot executes semantic
subgoals through four atomic skills: \textbf{Grasp}, \textbf{Place},
\textbf{Move\_to}, and \textbf{Observe\_state}.
We evaluate ProLoop on two representative tasks, breakfast preparation and tabletop
tidying, with 10 trials for each task. Across 20 trials, ProLoop achieves an overall
success rate of 90\%. Fig.~\ref{fig:real_world} illustrates a representative
six-subgoal breakfast preparation task, showing how progress assessment supports
continued execution, correction, and subgoal transition throughout the task.
Additional details and demonstrations of the real-world experiments are available
on our project website and in the supplementary videos.
\FloatBarrier

\section{Conclusion}

We present \textbf{ProLoop}, a progress-aware closed-loop framework for long-horizon robotic task planning.
ProLoop derives subgoal progress from structured relations between desired conditions and evolving execution states.
This progress guides timely closed-loop adaptation by preserving ongoing execution, triggering correction upon deviation, and permitting transition upon completion.
Completion-gated memory retains task-relevant information for subsequent planning.
Experiments on ALFWorld show improved task success with fewer environment interactions, while real-world trials demonstrate feasibility on a physical robot.
These results highlight the value of explicit progress assessment for reliable long-horizon execution.
%%%%%%%%%%%%%%%%%%%%%%%%%%%%%%%%%%%%%%%%%%%%%%%%%%%%%%%%%%%%%%%%%%%%%%%%%%%%%%%%
% arXiv preprint: references follow the conclusion directly (no \clearpage).
\bibliographystyle{IEEEtran}
\bibliography{references}

\end{document}